# Automated Alertness and Emotion Detection for Empathic Feedback During E-Learning

S L Happy[1], Anirban Dasgupta[2], Priyadarshi Patnaik[3], Aurobinda Routray[4]
[1,2,4]Department of Electrical Engineering, IIT Kharagpur, India
[3]Department of Humanities and Social Science, IIT Kharagpur, India

*Abstract*— In the context of education technology, empathic interaction with the user and feedback by the learning system using multiple inputs such as video, voice and text inputs is an important area of research. In this paper, a non-intrusive, standalone model for intelligent assessment of alertness and emotional state as well as generation of appropriate feedback has been proposed. Using the non-intrusive visual cues, the system classifies emotion and alertness state of the user, and provides appropriate feedback according to the detected cognitive state using facial expressions, ocular parameters, postures, and gestures. Assessment of alertness level using ocular parameters such as PERCLOS and saccadic parameters, emotional state from facial expression analysis, and detection of both relevant cognitive and emotional states from upper body gestures and postures has been proposed. Integration of such a system in e-learning environment is expected to enhance students' performance through interaction, feedback, and positive mood induction.

*Keywords- Affective learning; alertness assessment; emotion detection; intelligent e-learning system; multimodal recognition.*

## I. INTRODUCTION

In enhancing e-learning technology, human-computer-interaction plays an important role. In technology-supported classrooms, students are now armed with powerful tools to help them gather data and information, consult with peers, present their findings, and get up-to-date information from different sources with the use of computers that can assess their needs and orientations intelligently. Moreover, currently, research is underway to make such interactions affective as well. Apart from the use of voice and text commands, computers can now monitor students' emotion as well as activity through their voice, gesture, and facial expressions for more effective interaction and feedback during learning. Emotion, being an important element in learning, has gained attention of researchers for a better understanding of the role of affect in learning. Humans recognise emotional states by observing the visual and audio cues such as facial expressions, body posture, gesture, tone of voice etc. Therefore, it is easy for a classroom tutor to assess, inspire and motivate students, whereas in e-learning context the absence of emotional communication hinders the progress of the students. In order to resolve this issue, affective learning technology has been developed during the last decade which is a new multi-disciplinary area of research.

Recent findings suggest that when basic emotional mechanism are missing in the brain, then intelligent functioning is affected, while too much emotion is bad for rational thinking [1]. A slightly positive mood induces better feeling and the kind of thinking characterized by a tendency toward greater creativity in problem solving and decision making [2]. Affective states such as anger, sadness, fear, and happiness are associated with different patterns of blood flow explaining their pattern of influence on the brain. Research has demonstrated that positive emotions such as joy, acceptance, trust, and satisfaction can enhance learning, while prolonged emotional distress can demotivate the students and affect the learning process. Memorizing is affected by states of anxiety and depression, which often leads to despair, frustration, and worry, and is finally expressed through basic emotions like anger, sadness, and fear [3]. In such situations, intelligent feedback can support students and motivate them, thereby enhancing learning. For this purpose, the computer should be able to understand learners' emotions, interests, and attention and provide feedback to negotiate, dissipate or transform learners' emotions in an intelligent manner. In e-learning systems, intelligent and affective feedback can be provided through virtual personas such as embodied conversational agents (ECAs) or the embodied graphical agents that are capable of communicating with humans through verbal and nonverbal means and also by expressing emotions. This is considered more effective than a computer 'without a face' talking back. However, building such systems that recognize human emotion and alertness, and taking action accordingly is a challenging effort.

In this paper, we have put forward a model for automatic recognition of learners' cognitive state and providing appropriate interaction and feedback. The framework of affective state recognition involves the identification of basic emotions such as happiness, surprise, anger, sadness, fear, and sadness from the facial expression analysis. Further the emotions are classified to either positive or negative. The alertness level is estimated using PERcentage CLOSure of eye (PERCLOS) and saccadic parameters, the ocular measures of alertness level indicator. Similarly, the head tilt is also estimated from images. By combining these parameters, the system determines the appropriate feedback for the learner.

Section II describes the related literature in this fields and section III discusses the methodology followed to achieve the proposed goal. Finally, section IV discusses future work and section V concludes the paper.

## II. RELATED RESEARCH

The challenge of emotion recognition is associated with ambiguity and error while mapping emotional states to the factors that can be used to detect them. Work of Ekman reports the universal facial expressions [4] associated with basic emotions such as happiness, angry, sadness, surprise, fear and disgust. Research on automatic affect recognition received a boost after the late 1990s with success in affect recognition from face images and from audio-visual channels. Literature suggests ways of recognising emotions by observing the message displayed by the whole face or by observing the changes occurring in specific facial muscles, otherwise called as sign-judgment approach. By using Facial Action Coding System (FACS), facial actions are classified into different Action Units (AUs) and emotions are categorised using collection of AUs. Works of Bartlett [5] and Mattivi [6] report robust and correct detection of AUs leading to accurate emotion recognition. Geometric features based models trace the shape and size of the face and facial components such as the eye, mouth or the eyebrows and facial expression is categorized according to its variation. Holistic approaches use different machine learning approaches to extract features from the detected face and classify them into different emotional state. Using the developed applications, recognition of emotion from facial expressions from vision cue seems feasible. The FaceReader [7] uses FACS to distinguish six basic emotions with an accuracy of 89%. In [8], authors have achieved considerable accuracy in facial expression recognition by using local features in a person specific dataset. Facial expressions can also reveal the fatigue or boredom level in an individual though this is not that well researched.

Apart from the face, eyes can communicate if a person is bored or inattentive. Saccades are fast movements of both the eyes in the same direction. Saccades may be recorded using scleral search coils, EOG and high speed videos. Scleral coil and EOG based methods are contact based and hence infeasible for implementations in practical portable systems. Ueno *et al.* [9] have developed a system to assess human alertness and to alert the subject with acoustic stimulation on the basis of the dynamic characteristics of saccadic eye movement. In [10], authors have investigated the relationship between visual attention and saccadic eye movement. PERcentage CLOSure of eyes (PERCLOS) has been scientifically validated [11] as a generally useful and reliable index of lapses in visual attention using the well-known Psychomotor Vigilance Task (PVT). Combining the saccadic parameters along with the PERCLOS will increase the accuracy of alertness assessment. The work in [12] [13] describe about image based estimation of PERCLOS. An empathetic sotware Agent (ESA) is developed by Wang *et al.* [14] which interfaces eye movement information to assist empathy-relevant cognitive processes. Tracking eye gaze and pupil dilation in real-time, it can monitor the user's interest and attention, and personalize the agent's behaviors.

Moreover, gestures and postures also provide significant information about an individual's affective states, interest and attention level. However, these channels are not so well researched. Data from the above said channels of interactions can be combined to validate the cognitive state of the user.

Development of an affective tutoring systems mostly depends on the recognition of emotional state and an essential understanding of influences of emotion on learning, thereby providing appropriate feedback. A preliminary design of an architecture is proposed in [15] for including personality and emotional responsiveness in virtual tutoring agents. In [16], authors claim that the detection of frustration, boredom, confidence, motivation, and fatigue levels are necessary for a computer tutor and analyze the feedback type for each of the detected state. In [17], authors detect basic emotions like fear, anger, sadness, and happiness. They use ECAs to perform parallel empathy followed by reactive empathy by displaying emotion through expression and voice.

III. METHODOLOGY

The proposed learner monitoring system uses a webcam to monitor the activity of the user without any intervention. While assessment of emotion and alertness level from speech is also possible, our model proposes that there are at least four ways that visual data can be interpreted, thus creating a robust cognitive and affect detection system. Therefore, the proposed system entirely focuses on visual cues. From the face image, our system can estimate the alertness and attention level from ocular parameters as well as the emotional state from facial expressions. The recognition of posture and gesture can also be integrated for multimodal detection of cognitive state. For accurate classification of emotional state, we have used customized dataset for each user. The proposed model will fuse data from different visual channels of interaction such as facial expression, gesture, and eye parameters and assess the alertness level and emotional state of the learner in real-time. The integration of the estimated parameters provide appropriate feedback by feedback agent through audio, text, or haptic feedback. Within the scope of this paper, only brief descriptions of the algorithms and models are provided. The intended system architecture is presented in Fig. 1.

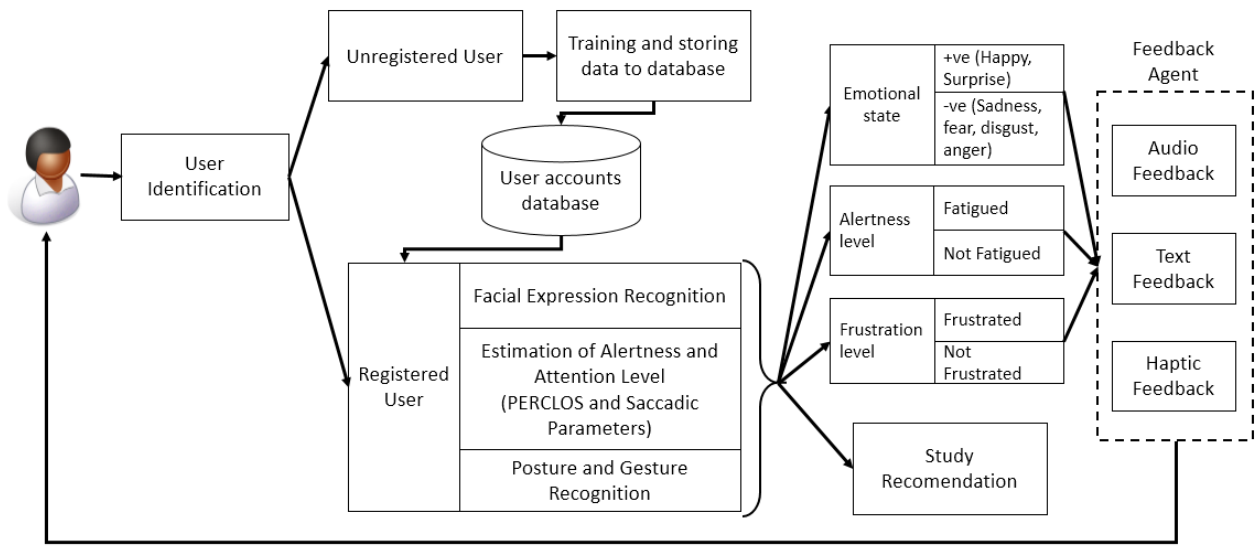

Fig. 1 Model for monitoring learners and proving feedback

### A. Detection of Emotional state

Towards achieving the goal of emotion recognition, the model detects face from the image and according to the extracted facial features, it classifies the expressions into different emotions. A real time customized algorithm has been developed that can classify facial expressions of a person into six basic emotions. PCA based dimensionality reduction has been used to reduce computational complexity. Local Binary Patterns (LBP) features, a local feature descriptor, have been used which takes care of minor changes of facial expressions for different emotions. The block LBP histogram features extract local as well as global features of face image resulting higher accuracy. The projection of different emotional facial images in feature space for a particular person is stored in the database which is further used while recognising emotions. After extraction of features from face images, different expressions are classified to different classes of emotions with considerable accuracy.

In this customized database, each user has to register and their emotional data is stored in the database. After user identification, the set of parameters of that user is fetched and used during emotion recognition process. The process of emotion recognition is shown in Fig. 2.

### B. Detection of Alertness level from ocular parameters

Measurement of PERCLOS is a standard for assessing the alertness level in an individual. PERCLOS is defined as the proportion of time that a subject's eyes are closed over a specified period. Similarly, eye saccade is important for measuring visual attention. Observing eye gaze, the interest of student from a set of options can be identified which will further helpful in adaptive hypermedia learning system. A real-time algorithm has been developed for estimating the PERCLOS value. Haar-like features have been used to detect the face. The potential region of finding the face is tracked using a Kalman Filter thereby reducing the search space for face detection and increasing the real-time performance. Affine and Perspective transformations have been used to consider the in-plane and off-plane rotations of the face respectively. A sample of rotated face detection is shown in Fig. 3. Eye detection is carried out using LBP based features. The eye state is classified as open or closed using SVM as shown in Fig. 4. A 66.67% overlapping window of three minutes duration [18] is selected to compute PERCLOS. Based on a threshold of 15%, the learner's alertness state is classified as drowsy or alert. A voice alarm is set to warn the learner, in the event of PERCLOS exceeding the threshold.

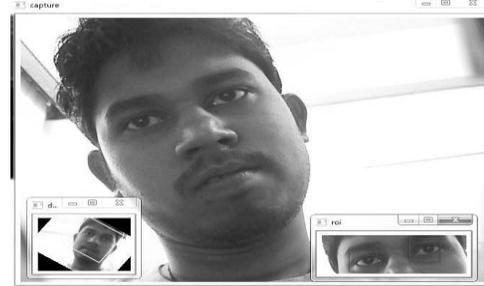

Fig. 3 Affine Transformation Based Detection of In-plane Rotated Faces

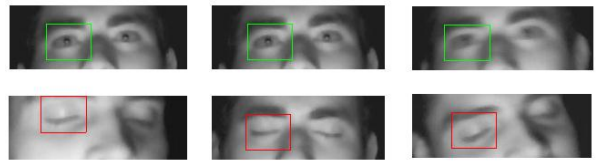

Fig. 4 Eye detection using LBP features and classification using SVM

The saccadic ratio (SR) estimation has also been carried out to indicate attention level. SR is the ratio of the peak saccadic velocity to the saccadic duration. The steps followed for saccadic parameter estimation are eye and eye corner detection followed by iris center estimation. Kalman filter has been implemented for accurately tracking the movement of eye.

### C. Recognising postures and gestures

Gestures as well as postures play significant role in understanding attention and interest of a learner. Gestures can be categorized into different activities through constant observation of body movement from vision cue. Similarly, postures can be identified from images which speaks intension and involvement of the user in learning process. The developed system uses affine transformation on the captured image to find out postures such as head tilt. Head tilt is reported as a good indicator of frustration [19]. Apart from this, posture of hand and head gestures also convey information for boredom and frustration level. Development

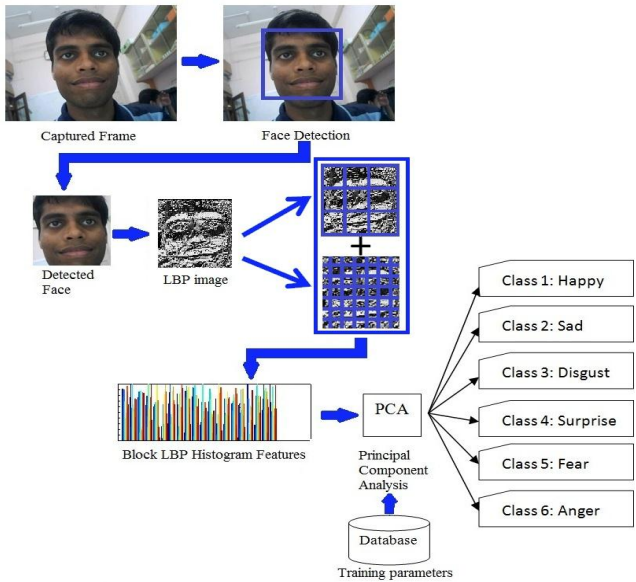

Fig. 2 Processing of the captured image to classify emotion

of algorithms to assess cognitive and emotional states using gestures is underway.

IV. FUTURE WORK

The estimation of affective state and assessment of attention level of the user has accomplished till date. The detected emotion is further divided into positive (happiness, surprise) and negative (sadness, fear, anger, disgust). Similarly the system detects positive and negative states for attention and frustration level as discussed earlier. The combination of these signals are provided as the input to the feedback system. The fusion of the sub-units for a complete assessment is our next objective. The design of the appropriate feedback for different emotional state and attention level is still under progress. In case of fatigue or frustration detection, audio feedback for taking a short break or refreshment through a video suggestion can be provided to relax the learner. Similarly, by haptic and text feedback, learners interest will be enhanced. During negative emotions, reactive emotional feedback will be provided to overcome the situation as well as to support the student. For recognition of different cognitive states such as boredom, alertness, and emotion from facial expression, postures and gestures, a database will be created and Electro-Encephalo-Graph data will be recorded for validation purpose. The gesture and posture analysis will be further carried out in case of boredom and frustration. The system can motivate for further learning and provide necessary material when positive state of mind is detected.

V. CONCLUSION

A model for intelligent assessment of alertness and emotional state in the context of learning and an appropriate feedback generation method has been proposed in this paper. The model proposes the use of nonintrusive visual cues to recognise emotion and alertness states using four diverse parameters – facial expression, ocular parameters, gestures, and postures – thereby having the potential to make very accurate judgment of cognitive and affect states using only a webcam. While most of such models and designs are integrated with specific e-learning systems, we propose a stand-alone system that can be used in different learning contexts independent of the e-learning system. This will help learners in a variety of e-learning environments and is expected to improve the productivity of learners through appropriate interaction and feedback.